\documentclass[letterpaper, 10 pt, conference]{ieeeconf}

\IEEEoverridecommandlockouts       
\makeatletter
\let\NAT@parse\undefined
\makeatother

\usepackage{hyperref}

\usepackage{amsmath} 
\usepackage{amssymb} 
\usepackage{graphicx}
\usepackage[table]{xcolor}
\usepackage{bm}
\usepackage{balance}
\usepackage{cite}
\usepackage{tabularx}
\usepackage{siunitx}
\usepackage{booktabs}
\usepackage{multirow}
\usepackage[font=footnotesize]{caption}

\usepackage[utf8]{inputenc}
\usepackage{pgfplots}
\DeclareUnicodeCharacter{2212}{−}
\usepgfplotslibrary{groupplots,dateplot}
\usetikzlibrary{patterns,shapes.arrows}
\pgfplotsset{compat=newest}

\hypersetup{
  colorlinks    = true, 
  urlcolor      = blue, 
  linkcolor     = blue, 
  citecolor     = red   
}
\newcommand{\RR}{\mathbb{R}}
\newcommand{\T}{^{\intercal}}
\newcommand{\hpz}{\hphantom{0}}

\newtheorem{definition}{Definition}

\newtheorem{assumption}{Assumption}

\newcommand\xqed[1]{%
  \leavevmode\unskip\penalty9999 \hbox{}\nobreak\hfill
  \quad\hbox{#1}}
\newcommand\qed{\xqed{$\blacksquare$}}
\newcommand\qeds{\tag*{\xqed{$\blacksquare$}}}

\definecolor{hlcolor}{rgb}{0, 0, 0}

\newcommand{\textref}[2]{\hyperref[#1]{#2~\ref*{#1}}}
\newcommand{\diffhl}[1]{\textcolor{hlcolor}{#1}}

\title{\LARGE \bf
Safety Filtering While Training: Improving the Performance and Sample Efficiency of Reinforcement Learning Agents}

\author{Federico Pizarro Bejarano, Lukas Brunke, and Angela P. Schoellig
\thanks{The authors are with the Learning Systems and Robotics Lab (\url{www.learnsyslab.org}), University of Toronto, Canada, and affiliated with the University of Toronto Robotics Institute and the Vector Institute for Artificial Intelligence in Toronto. Lukas Brunke and Angela P. Schoellig are also with the Technical University of Munich and the Munich Institute for Robotics and Machine Intelligence~(MIRMI), Germany. E-mails: \{federico.pizarrobejarano, lukas.brunke, angela.schoellig\}@robotics.utias.utoronto.ca}}

\begin{document}

\maketitle
\thispagestyle{empty}
\pagestyle{empty}

\begin{abstract}
Reinforcement learning~(RL) controllers are flexible and performant but rarely guarantee safety. Safety filters impart hard safety guarantees to RL controllers while maintaining flexibility. However, safety filters can cause undesired behaviours due to the separation between the controller and the safety filter, often degrading performance and robustness. In this paper, we analyze several modifications to incorporating the safety filter in training RL controllers rather than solely applying it during evaluation. The modifications allow the RL controller to learn to account for the safety filter, improving performance. This paper presents a comprehensive analysis of training RL with safety filters, featuring simulated and real-world experiments with a Crazyflie 2.0 drone. We examine how various training modifications and hyperparameters impact performance, sample efficiency, safety, and chattering. Our findings serve as a guide for practitioners and researchers focused on safety filters and safe RL.
\end{abstract}

\section{Introduction}
Robots are increasingly used for safety-critical applications such as autonomous driving~\cite{zeus} and surgery~\cite{surgery}. These tasks, characterized by complex cost functions and~(possibly unknown) dynamics, are challenging for classical controllers~\cite{brunke_safe_2021}. This motivates learning-based controllers, especially reinforcement learning~(RL) algorithms. Their ability to adapt to complex reward signals and unknown dynamics has led to superior performance in various domains~\cite{silver-nature-2016}. However, a significant limitation of RL is the lack of safety guarantees~\cite{brunke_safe_2021}. This is undesirable for deployment in safety-critical scenarios despite promising results.

\begin{figure}[t]
  \centering
  \includegraphics[width=0.95\linewidth]{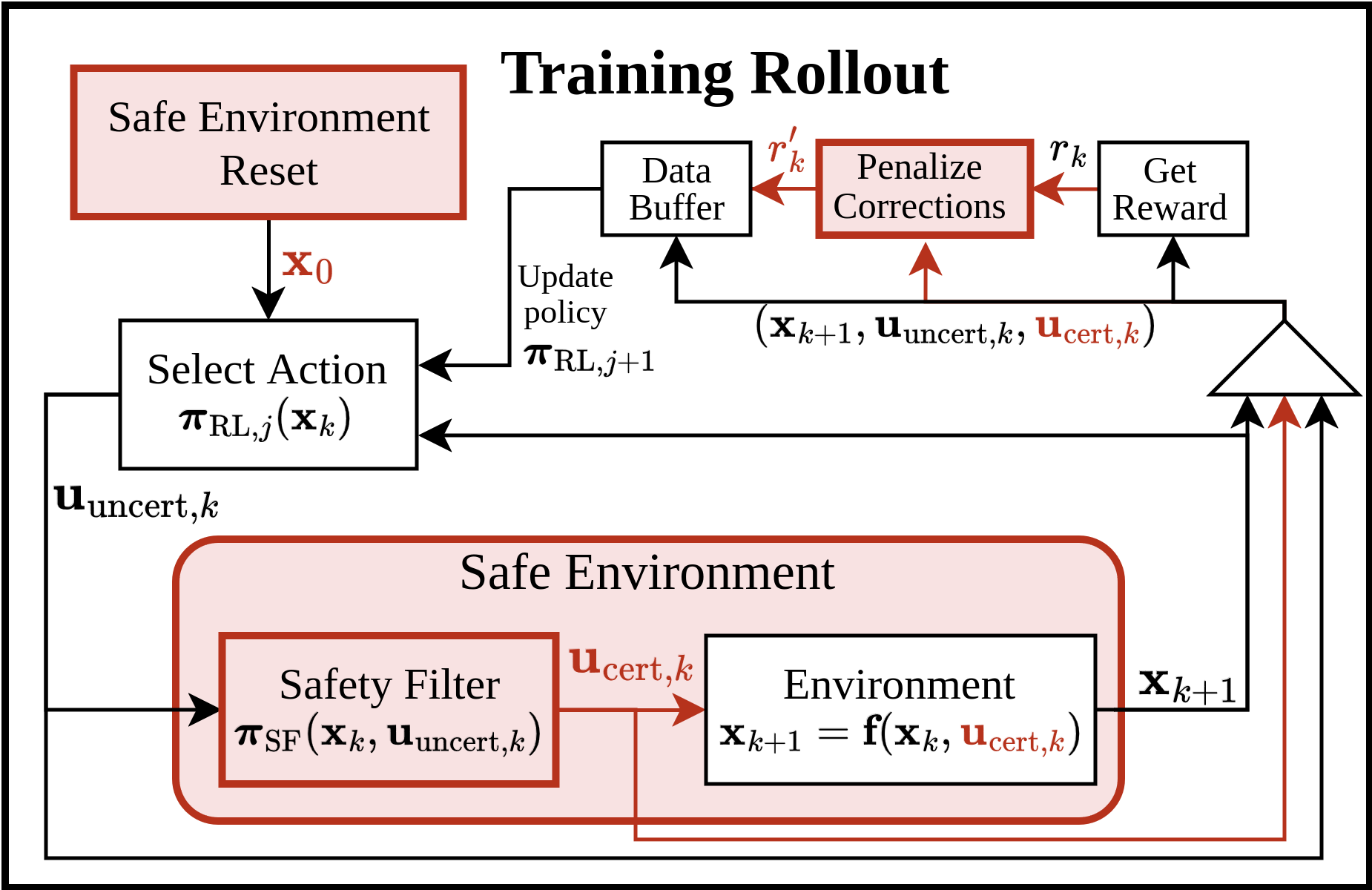}
  \caption{\diffhl{RL training with the safety filter modifications} in red. An RL agent takes in the current state $\textbf{x}_k$ and outputs a (potentially unsafe) action $\textbf{u}_{\text{uncert}, k}$. This action is passed to the safety filter, which outputs a minimally corrected safe action $\textbf{u}_{\text{cert}, k}$. This safe action is applied to the environment, and the difference between the safe and unsafe action is used to penalize the reward received by the RL agent. When an episode is completed, the environment is reset to a safe state using the safety filter.}
  \label{fig:summary_diagram}
  \vspace*{-7mm}
\end{figure} 

Safety filters can ensure that the RL controllers operate within defined constraints while minimally interfering with their operation. They determine whether uncertified (i.e., potentially unsafe) inputs from the controller will violate the constraints~\cite{fisac_survey, zeilinger_survey}. If so, the filter determines the minimal deviation from the input that results in constraint satisfaction.

However, adding a safety filter changes how the controller interacts with the environment. By incorporating the safety filter in training, the RL algorithm can train on the system on which it will be evaluated (i.e., the system certified by the safety filter)~\diffhl{\cite{hanna_survey}}. This way, it can learn to account for the safety filter, maximizing rewards on the desired system. Additionally, using the safety filter during training can eliminate constraint violations during training, allowing RL controllers to be trained directly on physical systems. Finally, adding the safety filter significantly improves convergence as the RL controller will not explore unsafe states or cause complete failures (e.g., a drone crashing) and instead focuses on the desired safe states~\diffhl{\cite{hanna_survey}}.

\textit{Contributions}: In this paper, we \diffhl{analyze} three modifications to the training process of any RL controller through the incorporation of a safety filter (see \textref{fig:summary_diagram}{Fig.}). We use model predictive safety filters~(MPSFs), but the \diffhl{training} modifications can be equally applied with other types of safety filters. The training \diffhl{modifications} are demonstrated in simulation using the \texttt{safe-control-gym}~\cite{safe-control-gym} on a quadrotor system and on a real-world Crazyflie 2.0 quadrotor (see \textref{fig:drone_chat}{Fig.}). We found that \diffhl{the} modifications significantly improve sample efficiency, eliminate constraint violations during training, and improve final performance \diffhl{and reduce chattering~\cite{multi-step}} on the certified system. Our code can be found at \url{https://tinyurl.com/sf-train-code}, and a video of our experiment can be found at \url{https://tinyurl.com/sf-train-video}.

\section{Related Work}
\subsection{Safety Filters} \label{sec:rel_w_sf}
Safety filters can provide any controller with hard safety guarantees by determining the minimal input adjustments~(corrections) necessary to ensure constraint satisfaction~\cite{fisac_survey, zeilinger_survey}. Control barrier functions~(CBFs) identify a safe set as the superlevel set of a function and guarantee safety through an online optimization that keeps the system in the safe set~\cite{cbfs}. Generating CBFs requires knowledge of the system dynamics or offline data, they are typically defined in continuous time, and they do not guarantee input-constraint satisfaction~\cite{brunke_safe_2021}. Other methods of defining a safe set, such as Hamilton-Jacobi reachability analysis~\cite{zeilinger_survey}, also provide hard safety guarantees but can be equally difficult or computationally expensive to determine. Model predictive safety filters~(MPSFs) leverage model predictive control~(MPC) to assess the safety of proposed inputs~\diffhl{\cite{zeilinger_survey, zeilinger_nonlinear}}. By simulating input trajectories using a system dynamics model over an MPC horizon, the MPSF can guarantee that a safe trajectory exists from the current state to a safe terminal set, guaranteeing safety and recursive feasibility. MPSFs tend to have higher online computation demands than other safety filters but require less offline computation.

Controllers are typically designed without consideration for the safety filter and thus will not choose optimal actions for the certified system. This may result in diminished performance and chattering~\cite{zeilinger_survey, multi-step}, where the controller persistently attempts to violate constraints, only to be continually stopped by the safety filter, resulting in oscillatory, inefficient behaviour. 

\subsection{Reinforcement Learning}
Reinforcement learning~(RL) is a framework for sequential decision-making that aims to find a policy that maximizes the sum of attained rewards. RL typically assumes that the underlying dynamics function and reward signal are unknown. RL's lack of explicit assumptions and constraints limits safety but promotes expressiveness~\cite{sutton_and_barto, brunke_safe_2021}. 

Several strategies have been proposed to develop controllers with the expressiveness offered by RL while providing safety guarantees. We will focus on those that aim toward state and input constraint satisfaction.

One approach is to use a safety layer to convert an optimal but potentially unsafe action produced by the RL policy to the closest safe action with respect to the state constraints~\cite{safe-explorer, safe-explorer-2}, which is similar to the safety filters discussed in \textref{sec:rel_w_sf}{Section}. However, the optimization only attempts to prevent constraint violations in the next time step rather than guaranteeing safety for all future times.

Constrained Markov decision processes~(CMDPs)~\cite{cmdp} extend Markov decision processes (MDPs) with constraints. However, most of these approaches suffer from significant computational complexity, confining demonstrations to naive tasks~\cite{brunke_safe_2021}. One approach to solve the CMDP problem is to transform the problem into an unconstrained optimization using Lagrangian methods and use RL as a subroutine in the primal-dual updates~\cite{cpo}.

While these methods encourage safety, they do not have the hard guarantees often required in safety-critical applications~\cite{brunke_safe_2021, safe_rl_w_constr_admis_set}. The absence of a model of the system also tends to limit the sample efficiency of these methods~\cite{brunke_safe_2021, end2end-safety}. Additionally, these methods do not prevent constraint violations during training and may need to violate the constraints to learn a safe policy~\cite{cpo}, making them unsuitable for training RL controllers directly on physical systems. 

\subsection{Safety Filtering During Training}
Filtering the actions of an RL agent during training has been approached using several methods~\diffhl{\cite{hanna_survey}}. State-wise safe sets are used in~\cite{safe_rl_w_constr_admis_set, zonotopes, learn_w_imagination}, CBFs are used in~\diffhl{\cite{end2end-safety, robust_cbf, wang_car}}, and~\diffhl{\cite{zeilinger_nonlinear, diesel, satellite} use MPSFs}. To encourage the RL agent to pursue safe actions, the reward signal \diffhl{has been} penalized by a \diffhl{constant} penalty~\diffhl{\cite{satellite, hanna_survey}}, \diffhl{a penalty proportional to the correction~\cite{zeilinger_nonlinear, wang_car}, or a filter-specific penalty~\cite{chance-constr}}. Differentiable safety layers maintain safety in training and evaluation while jointly optimizing the controller and safety filter~\cite{barriernet, diff_cbf}, but require additional assumptions such as perfect system knowledge. \diffhl{Training with a safety filter has been found to} improve sample efficiency as the RL agent only explores safe states and occasionally improve the final performance~\diffhl{\cite{zonotopes, diesel, hanna_survey}}.

\diffhl{This paper uses} model predictive safety filters, which do not require an explicit representation of the safe set. However, the modifications described in \textref{sec:modifications}{Section} can also be applied to other safety filters~\diffhl{\cite{hanna_survey}}. \diffhl{Unlike approaches that aim to phase out the safety filter during training~\cite{satellite, diesel}, our focus is on the final \textit{certified} performance. We examine the impact of each modification individually and in combination, alongside an analysis of how varying the correction penalty weight compares to constraint violation penalties~\cite{zonotopes, reward_shaping}. This work aims to compile, formally describe, and analyze training modifications with a safety filter, providing a resource} on including safety filters into any RL training procedure.

\section{Problem Formulation}
We consider a discrete, time-invariant system given by
\begin{align} \label{eq:true-system}
    \textbf{x}_{k+1} = \textbf{f}(\textbf{x}_{k}, \textbf{u}_k),
\end{align}
where $\textbf{x}_k \in \mathbb{X}$ represents the system state at time step $k$, $\textbf{u}_k \in \mathbb{U}$ denotes the control input, and $\textbf{f}$ encapsulates the system dynamics. The system is subject to known state and input constraints $\textbf{x} \in \mathbb{X}_{\text{c}}, \;\textbf{u} \in \mathbb{U}_{\text{c}}$, where $\mathbb{X}_{\text{c}} \subset \mathbb{X} \subset \RR^n$ is closed and $\mathbb{U}_{\text{c}} \subset \mathbb{U} \subset \RR^m$ is compact.

We assume that we only have access to a nominal model~$\bar{\textbf{f}}$ and compact uncertainty set $\mathbb{W}$ such that 
\begin{align} \label{eq:nominal-system}
    \textbf{x}_{k+1} = \bar{\textbf{f}}(\textbf{x}_{k}, \textbf{u}_k) + \textbf{w}(\textbf{x}_k, \textbf{u}_k) \,,
\end{align}
where $\textbf{w}(\textbf{x}, \textbf{u}) \in \mathbb{W} \subset \RR^n \,, \forall \textbf{x} \in \mathbb{X}_{\text{c}}\,, \forall \textbf{u} \in \mathbb{U}_{\text{c}}$.

A safety filter, which has access to the nominal model $\bar{\textbf{f}}$, takes in the current state of the system $\textbf{x}_{k}$ and input $\textbf{u}_{\text{uncert}, k} = \pi_{\text{uncert}}(\textbf{x}_k)$, and uses the nominal system $\bar{\textbf{f}}$ and knowledge of the uncertainty set $\mathbb{W}$ to determine if the proposed input is safe~(i.e., satisfies state and input constraints, and will not lead to a constraint violation in the future). If it is not safe, it will find a safe input $\textbf{u}_{\text{cert}, k}$ that minimally modifies the uncertified input. This certified input is the input applied to the system (i.e., $\textbf{u}_k = \textbf{u}_{\text{cert}, k}$). 

\section{Background}
\subsection{Model Predictive Safety Filters} \label{sec:back_MPSF}
MPSFs can be based on any MPC framework, inheriting the safety and recursive feasibility guarantees of the underlying MPC~\cite{multi-step}. A general formulation of MPSFs based on robust tube-based MPC is provided below to motivate MPSFs. The specific implementation used for experiments is the robust nonlinear MPC described in~\cite{nl_mpc}. This is further elaborated on in \textref{sec:experiments}{Section}.

\begin{definition}[Robust pos. control inv.~(RPCI) set~\cite{mpc_textbook}] \label{sec:RPCI}
A set $\mathbb{P} \subseteq \mathbb{X}_{\text{c}}$ is robust positively control invariant~(RPCI) for the system in~\textref{eq:nominal-system}{Eq.} with a controller $\pi_{\text{RPCI}} : \mathbb{P} \to \mathbb{U}_{\text{c}}$ if $\forall \textbf{x} \in \mathbb{P}$ and $\forall \textbf{w} \in \mathbb{W}$:
\begin{align*}
    & \bar{\textbf{f}}(\textbf{x}, \pi_{\text{RPCI}}(\textbf{x})) + \textbf{w} \in \mathbb{P}. \qeds
\end{align*}
\end{definition}

\vspace{2mm}
\begin{assumption} \label{assump:1}
    There exists a terminal set $\mathbb{X}_{\text{term}} \subset \mathbb{X}_{\text{c}}$ and a terminal controller $\pi_{\text{term}} : \mathbb{X}_{\text{term}} \to \mathbb{U}_{\text{c}}$ such that $\mathbb{X}_{\text{term}}$ is a RPCI set for the system in~\textref{eq:nominal-system}{Eq.} under $\pi_{\text{term}}$. \qed
\end{assumption}

An MPSF solves an optimization problem at each time step for an optimal input sequence over the next $H$ time steps, where $H \in \mathbb{N}$ is the horizon. At each time step $k$, the MPSF only applies the first input from the resulting input sequence. The optimization problem for a general robust, tube-based multi-step MPSF can be stated as follows~\cite{multi-step}:
\begin{subequations}
\begin{align}
    \min_{\textbf{u}_{\cdot|k},\, \mathbb{X}_{\cdot|k}} & \sum_{j=0}^{M-1} w(j)\| \pi_{\text{uncert}}(\textbf{z}_{j|k}) - \textbf{u}_{j|k} \|^2_2 \\
    \text{s.t. } & \textbf{x}_k \in \mathbb{X}_{0|k},\, \textbf{z}_{0|k} = \textbf{x}_k, \label{eq:nominal-mpc-init} \\ 
    &\mathbb{X}_{i+1|k} \supseteq \boldsymbol{\Phi}(\mathbb{X}_{i|k}, \textbf{u}_{i|k}, \mathbb{W}), \\
    & \textbf{z}_{i+1|k} = \bar{\textbf{f}}(\textbf{z}_{i|k}, \textbf{u}_{i|k}), \\
    & \mathbb{X}_{i|k} \subseteq \mathbb{X}_{\text{c}}, \\
    & \textbf{u}_{i|k} \in \mathbb{U}_{\text{c}} \,, \; \forall \; i = 0, ..., H-1, \\
    & \mathbb{X}_{H|k} \subset \mathbb{X}_{\text{term}} \label{eq:nominal-mpc-terminal},
\end{align}
\end{subequations}
where $\textbf{u}_{i|k}$ is the input at the $(k+i)\text{-th}$ time step computed at time step~$k$, $M$ is the filtering horizon, $w(\cdot) : \mathbb{N}_{0} \to \RR^+$ calculates the weights associated with the $j\text{-th}$ correction, $\pi_{\text{uncert}}$ is the uncertified controller, $\textbf{z}_{j|k}$ is the estimated future state at the $(k+j)\text{-th}$ time step computed at time step~$k$, $\mathbb{X}_{i|k}$ is the set of possible states at the $(k+i)\text{-th}$ time step computed at time step $k$, and the evolution of the system is $\mathbb{X}_{i+1|k}  \supseteq \boldsymbol{\Phi}(\mathbb{X}_{i|k}, \textbf{u}_{i|k}, \mathbb{W}) = \{\bar{\textbf{f}}(\textbf{x}, \textbf{u}_{i|k}) + \textbf{w} \mid \forall\, \textbf{x} \in \mathbb{X}_{i|k}, \textbf{w} \in \mathbb{W}\}$. The future actions of the controller $\pi_{\text{uncert}}(\textbf{z}_{j|k})$ are approximated using~\cite{multi-step}.

In~\cite{multi-step}, the \textit{norm of the rate of change of the inputs} metric is proposed. This measures how much the input varied during an experiment. Consider the applied inputs $\textbf{u}_{k}$ for $k = 0, ..., K-1$. First, we take the numerical derivative $\delta \textbf{u}_{k} = (\textbf{u}_{k} - \textbf{u}_{k-1})/\delta t$ for $k = 1, ..., K-1$ (where $\delta t$ is the length of a time step) and stack them into a matrix $\Delta \textbf{u} = [\delta \textbf{u}_{1}, ..., \delta \textbf{u}_{K-1}]$. Then the norm of the rate of change of the inputs is $\|\Delta \textbf{u}\|_{\text{F}}$, where $\|\cdot\|_{\text{F}}$ is the Frobenius norm. We will use this metric to evaluate chattering and jerkiness.

\subsection{Training Reinforcement Learning Algorithms}
RL algorithms learn a policy to maximize the accumulated rewards by interacting with the environment. RL assumes the control problem is a Markov decision process~(MDP). An MDP comprises a state space $\mathbb{X}$, an input (action) space $\mathbb{U}$, a set of starting states $\mathbb{S} \subset \mathbb{X}$, a transition model $\mathcal{T} : \mathbb{X} \times \mathbb{U} \rightarrow \mathbb{X}$, and a per-step reward function $\mathcal{R} : \mathbb{X} \times \mathbb{U} \rightarrow \mathbb{R}$%
~\cite{sutton_and_barto}. RL algorithms typically assume the dynamics $\mathcal{T}$ and reward function $\mathcal{R}$ are unknown~\cite{brunke_safe_2021}. 

RL aims to find the optimal policy $\pi^\star: \mathbb{X} \rightarrow \mathbb{U}$ which maximizes the discounted accumulated reward (i.e., the return) from any starting state $\textbf{x}_0 \in \mathbb{S}$~\cite{sutton_and_barto, safe_rl_w_constr_admis_set}:
\begin{align}
    G_{\pi}(\textbf{x}_0) &= \mathcal{R}(\textbf{x}_0, \pi(\textbf{x}_0)) + \gamma \mathcal{R}(\textbf{x}_1, \pi(\textbf{x}_1)) + \dots \\
    &= \sum_{k=0}^\infty \gamma^k \mathcal{R}(\textbf{x}_k, \pi(\textbf{x}_k)),
\end{align}
where $G_{\pi}(\textbf{x}_0)$ is the return starting from state $\textbf{x}_0$ under the policy $\pi$, $\gamma \in (0, 1]$ is a discount factor, and \diffhl{$\textbf{x}_{k+1} = \mathcal{T}(\textbf{x}_{k}, \pi(\textbf{x}_{k})), \; \forall k \ge 0$}.

The agent improves the policy by interacting with the environment. At each time step $k$, the agent receives an observation $\textbf{x}_k \in \mathbb{X}$, takes an action $\textbf{u}_k \in \mathbb{U}$, transitions to a next state $\textbf{x}_{k+1} = \mathcal{T}(\textbf{x}_k, \textbf{u}_k)$, and receives a reward $r_k = \mathcal{R}(\textbf{x}_k, \textbf{u}_k)$. These four values form a quadruple $(\textbf{x}_k, \textbf{u}_k, \textbf{x}_{k+1}, r_k)$, which is used to improve the policy. On-policy RL algorithms require the quadruples to be from the current policy. In contrast, off-policy algorithms can learn from data generated by other policies~\cite{sutton_and_barto}. 

Training is typically divided into episodes, where a starting state $\textbf{x}_0$ is sampled from $\mathbb{S}$. The policy controls the system for a number of steps up to a maximum $K$, or until another condition is met~\cite{sutton_and_barto}. Then, the system is reset to a new starting point for the next episode.  

\section{\diffhl{Training Modifications}} \label{sec:modifications}
We consider three modifications to the training of RL algorithms~\diffhl{\cite{hanna_survey}}. These can be combined or used separately and can be applied to any RL controller and any safety filter, not only MPSFs.

\subsection{Filtering Training Actions} \label{sec:filter_train_actions}
During RL training, the controller generates uncertified actions $\textbf{u}_{\text{uncert}, k} \in \mathbb{U}$. By applying the safety filter $\textbf{u}_{\text{cert}, k} = \pi_{\text{SF}}(\textbf{x}_{k}, \textbf{u}_{\text{uncert}, k})$ to certify these actions, safety is guaranteed during training. 

For on-policy algorithms, the uncertified action generated by the RL controller must be the one buffered for policy improvement. Thus the buffered quadruple is $(\textbf{x}_k, \textbf{u}_{\text{uncert}, k}, \textbf{f}(\textbf{x}_k, \textbf{u}_{\text{cert}, k}), \mathcal{R}(\textbf{x}_k, \textbf{u}_{\text{cert}, k}))$~\diffhl{\cite{hanna_survey}}. Effectively, the RL algorithm operates in a safe environment protected by the safety filter rather than in the true, unsafe environment. For off-policy algorithms, this same approach can be done, or the safety filter can be considered an expert, and thus the safe action $\textbf{u}_{\text{cert}, k}$ can be buffered rather than the unsafe action. \diffhl{A detailed comparison of the learning quadruples and their effect on training can be found in~\cite{hanna_survey}}.

This modification aims to maximize the reward on the final certified system, prevent violations during training (assuming that the episodes start from safe states, see \textref{sec:safe_reset}{Section}), and improve sample efficiency by focusing the training on the safe areas. However, the RL agent may grow dependent on the safety filter and thus not perform well without the safety filter~\diffhl{\cite{satellite, safe_rl_w_constr_admis_set, wang_car}}. This is not a problem if the safety filter is used during evaluation, as assumed in this paper, and can be mitigated using the following approach~\diffhl{\cite{wang_car}}.

\subsection{Penalizing Corrections}
Penalizing constraint violations is a common approach to encouraging safety in RL~\cite{reward_shaping}. This is typically done by introducing a \diffhl{constant} penalty $\beta \ge 0$ to the reward when a constraint is violated~\diffhl{\cite{satellite, zonotopes}}, or replacing the reward entirely with the negative penalty $-\beta$~\cite{safe_rl_w_constr_admis_set}. With the safety filter, we can penalize corrections rather than constraint violations. Corrections indicate that the proposed action was unsafe before a constraint violation, and the magnitude of the correction measures how unsafe the action was. 

In~\diffhl{\cite{satellite, hanna_survey}}, the reward \diffhl{is} penalized by a \diffhl{constant penalty} when the safety filter \diffhl{is} activated. However, this does not incorporate how unsafe the proposed action \diffhl{is}. \diffhl{In~\cite{zeilinger_nonlinear, wang_car}, the reward is penalized by the magnitude of the correction. Filter-specific penalties have been proposed~\cite{chance-constr} but cannot be applied with other safety filters. The} general framework for utilizing the safety filter to penalize the reward \diffhl{is}:
\begin{align*}
    &\mathcal{R}^{\alpha}(\textbf{x}_k, \textbf{u}_{\text{uncert}, k}, \textbf{u}_{\text{cert}, k}) \\
    &= \mathcal{R}(\textbf{x}_{k}, \textbf{u}_{\text{applied}, k}) - \alpha \mathcal{P}(\textbf{u}_{\text{uncert}, k}, \textbf{u}_{\text{cert}, k}),
\end{align*}
where $\alpha$ weighs the correction penalty, $\mathcal{P}$ defines the penalty based on the correction, and $\textbf{u}_{\text{applied}, k}$ is the action applied to the system ($\textbf{u}_{\text{uncert}, k}$ in standard RL training, and $\textbf{u}_{\text{cert}, k}$ if filtering the training actions as discussed in the previous section). After experimentation, we found that penalizing the magnitude of the correction $\mathcal{P} = \|\textbf{u}_{\text{uncert}, k} - \textbf{u}_{\text{cert}, k} \|_2^2$~\diffhl{\cite{zeilinger_nonlinear, wang_car}} led to the best results in our experiments. 

This modification encourages safety and minimizes corrections but does not enforce safety during training or maximize the true reward. This reward-shaping modification should converge to an optimal, safe solution if combined with the previous section's filtering training actions modification. This is because an optimal RL policy will never propose an action that causes a correction (as it would be penalized and corrected); thus, every action will be safe. However, in practice, the weight $\alpha$ will affect the behaviour of the resulting controller, and there are no guarantees that the converged policy will always produce safe actions~\diffhl{\cite{sutton_and_barto, wang_car}}.

\subsection{Safely Resetting the Environment} \label{sec:safe_reset}
Sample efficiency can be improved by using the safety filter to avoid initiating an episode in an unsafe state~\diffhl{\cite{safe_reset}}. Often, there is no prescribed starting state, and thus $\mathbb{S} := \mathbb{X}_{\text{c}}$. However, even if $\textbf{x}_0 \in \mathbb{X}_{\text{c}}$, this does not guarantee that there exists an input trajectory that will keep the system safe for all future time. Using the safety filter, safety can be guaranteed by restricting $\mathbb{S}$ to a $H$-step robust positively control invariant~($H$-RPCI) set~\diffhl{\cite{safe_reset}}. 

An $H$-RPCI set guarantees that the system can be safely driven to a RPCI set (see \textref{sec:RPCI}{Definition}) in $H$ steps. We wish to find the $H$-RPCI set that can be driven to the terminal set $\mathbb{X}_{\text{term}}$. \diffhl{The $H$-RPCI set can be explicitly computed and sampled from, but this can be difficult~\cite{RPCI}}. Instead, we will sample $\textbf{x}_0 \sim \mathbb{S}$, then determine the feasibility of certifying an input from that state~\diffhl{\cite{safe_reset}}. If the safety filtering optimization is feasible, $\textbf{x}_0$ is within the $H$-RPCI set. If infeasible, another starting state is randomly generated until a feasible starting state is found.

This modification ensures that episodes only commence from certifiably safe states, improving sample efficiency as the RL training focuses on feasible trajectories~\diffhl{\cite{safe_reset}}. When combined with filtering training actions (\textref{sec:filter_train_actions}{Section}), this ensures the constraints are never violated during training.

\section{Experimental Results} \label{sec:experiments}
To determine the efficacy of \diffhl{the training modifications}, we ran experiments in the safe learning-based control simulation environment \texttt{safe-control-gym}~\cite{safe-control-gym} and on a real quadrotor, the Crazyflie 2.0 (see \textref{fig:drone_chat}{Fig.}). The underlying MPC is the robust nonlinear MPC detailed in~\cite{nl_mpc}, which assumes that the system is incrementally stabilizable and that the model mismatch between the nominal system and the real system is contained within a known compact set $\mathbb{W}$. The upper bound on the model mismatch \diffhl{$w_{\text{max}} = \max_{k \in \{0, \dots, K-1\}} \|\textbf{x}_{k+1} - \textbf{z}_{1|k}\|_2$} was found experimentally by comparing the true states \diffhl{$\textbf{x}_{k+1}$} and the states predicted by the nominal model \diffhl{$\textbf{z}_{1|k}$} while executing trajectories for system identification (further described below). We define $\mathbb{W} = \{\textbf{w}\in\RR^n \mid \|\textbf{w}\|_2 \le w_{\text{max}}\}$. The terminal set is determined from the uncertainty set $\mathbb{W}$ and a hand-tuned LQR terminal controller $\pi_{\text{term}}$~\cite{nl_mpc}. We use \texttt{acados}~\cite{acados} to efficiently solve the MPC optimization at each time step.

\begin{figure}[t]
  \centering
  \vspace{2mm}
  \includegraphics[width=1.0\linewidth,trim={30 40 50 25}, clip]{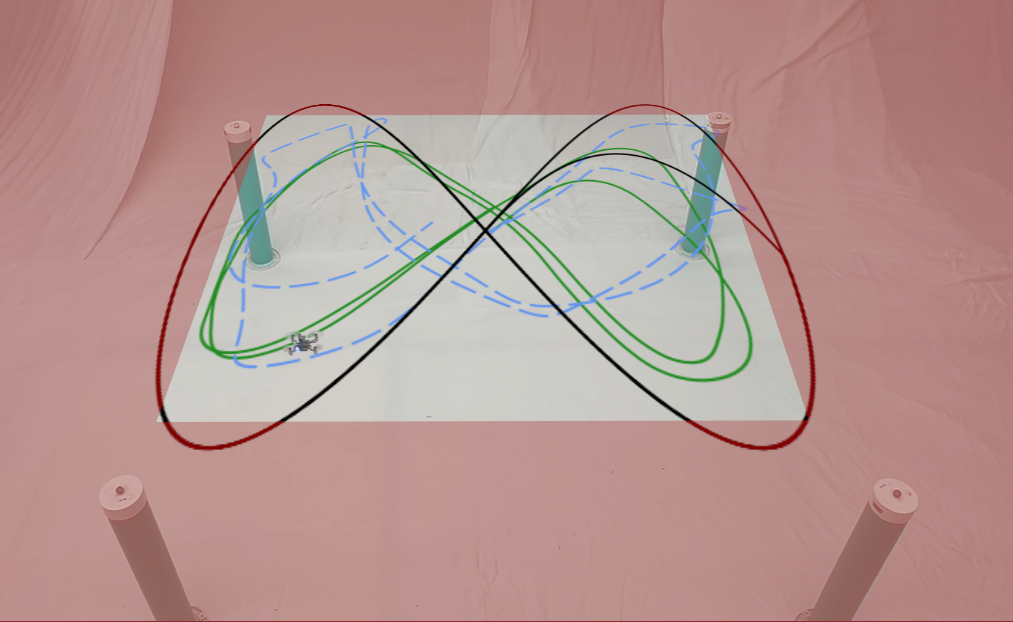}
  \caption{Experimental setup on a Crazyflie 2.0 drone. A PPO controller trained without a safety filter or constraint violation penalization (blue) tries to track a reference trajectory (black), but unforeseen interactions with the safety filter cause poor tracking performance. However, when trained with a safety filter~(green, with $\alpha=1$), the behaviour is smoother and more performant. The constraints are shown in red.}
  \label{fig:drone_chat}
  \vspace{-5mm}
\end{figure}

During training, we do not use early stopping (i.e., ending an episode when a constraint is violated as used in~\cite{safe_rl_w_constr_admis_set}), as it was found to prevent convergence when not filtering actions. The MPSF has a filtering horizon $M=1$ during training and $M=2$ during evaluation, which was found to reduce chattering in all approaches while minimally increasing computation. The state constraints were softened to allow for feasibility at all time steps, which was found to improve simulation results and allow for the successful execution of real-world experiments. This was done by incorporating slack variables and penalizing their magnitude with linear and quadratic terms.

Proximal policy optimization~(PPO)~\cite{ppo} was used as the RL controller due to its popularity in robotics tasks~\cite{safe-control-gym}. Each approach is trained five times with different seeds to determine the variability in the results. Safe explorer~\cite{safe-explorer} and constrained policy optimization~\cite{cpo}, two popular safe model-free RL approaches, were tested and found to not converge for performed tasks, likely due to the tasks' high dimensionality and difficulty in ensuring safety.

The reward signal for the RL algorithm at time step $k$ is $r_k = \exp(-2\|\textbf{p}_k - \textbf{p}_{\text{ref},k}\|_2^2)$, where $\textbf{p}_k$ is the current position of the drone and $\textbf{p}_{\text{ref},k}$ is the reference position at time step $k$. During training, we use two modified reward signals. For the baseline approaches we penalize constraint violations by a \diffhl{constant} penalty, $r_k^{\beta} = r_k - \textbf{1}_{\text{viol}}\beta$, where $\textbf{1}_{\text{viol}}$ is an indicator function for violations, equal to 1 when a violation occurs and is otherwise zero, and $\beta \ge 0$ is the constraint violation penalty. For \diffhl{the} approaches which penalize corrections, we use $r^{\alpha}_k = r_k - \alpha \|\textbf{u}_{\text{uncert}, k} - \textbf{u}_{\text{cert}, k} \|_2^2$~\diffhl{\cite{zeilinger_nonlinear, wang_car}}, where $\alpha$ is the correction penalization weight. The maximum reward achievable is $r_k = r^{\alpha}_k = r^{\beta}_k = 1$, for a total undiscounted return equal to the maximum number of iterations. The return in this section always refers to the undiscounted return~(i.e., \diffhl{$\gamma = 1$}) without using any penalties~(i.e., using $r_k$ as the reward).

\subsection{Ablation Study of the Training Modifications} \label{sec:ablation}
Every combination of the modifications was trained and evaluated to analyze the effects of \diffhl{the} training modifications. Additionally, some combinations are similar to the related works that only filter actions~(e.g.,~\cite{safe_rl_w_constr_admis_set, robust_cbf, diesel}) and filter actions as well as penalize corrections~\cite{chance-constr, satellite}; thus, this ablation study also serves to better analyze the related works. In the following section ``Std. ($\beta$$=$0)'' refers to the baseline with none of \diffhl{the} training modifications nor constraint violation penalties. The other approaches are combinations of \diffhl{the} training modifications: FA = Filtering Actions, PC = Penalizing Corrections, SR = Safe Reset.

The controllers were evaluated on a simulation of a Crazyflie 2.0 using the \texttt{safe-control-gym}~\cite{safe-control-gym}. The state is $\textbf{x} = [x, \dot{x}, y, \dot{y}, z, \dot{z}, \phi, \theta, \psi, \dot{\phi}, \dot{\theta}, \dot{\psi}]\T$, where the $x-y-z$ values are the position, $\dot{x}-\dot{y}-\dot{z}$ are the velocities, $\phi-\theta-\psi$ are the roll-pitch-yaw angles, and $\dot{\phi}-\dot{\theta}-\dot{\psi}$ are the body rates of those angles. The inputs are the thrusts applied to the four rotors. The nominal equations of motion can be found in~\cite{carlos}. These equations were discretized to get the nominal discrete model of the 3D quadrotor, which was very accurate, with $w_{\text{max}}=0.0205$. This uncertainty was determined by starting at random states in $\mathbb{X}_{\text{c}}$, executing random actions in $\mathbb{U}_{\text{c}}$, and measuring the model mismatch. The MPSF had an MPC horizon of $H=20$ and a frequency of \SI{50}{\hertz}. \diffhl{The} correction penalization weight is set to $\alpha = 1$ when using the correction penalization approach. \diffhl{The} constraint violation penalty is set to $\beta = 0$ \diffhl{in all cases}.

The trajectory tracking task consists of tracking a figure-eight reference with an amplitude of \SI{2}{\meter} in the $x$ dimension, \SI{0.5}{\meter} in the $y$ dimension, \SI{1}{\meter} in the $z$ dimension, which must be flown once in \SI{5}{\second}. The total number of iterations, and thus the maximum return, is \SI{50}{\hertz} $\cdot$ \SI{5}{\second} $= 250$. The position is constrained to be 5\% smaller than the full extent of the trajectory, the velocities are constrained to be within [\SI{-2}{\meter\per\second}, \SI{2}{\meter\per\second}], the angles were constrained to be within [\SI{-0.5}{\radian}, \SI{0.5}{\radian}], and the angular rates were constrained to be within [\SI{-2}{\radian\per\second}, \SI{2}{\radian\per\second}].

To find suitable starting points for evaluating the final trained models, certifiably safe states were found using the MPSF (using the process described in \textref{sec:safe_reset}{Section}). One hundred starting points were generated, and each algorithm was tested using these starting states. The starting states and the different training seeds are the source of randomness. 


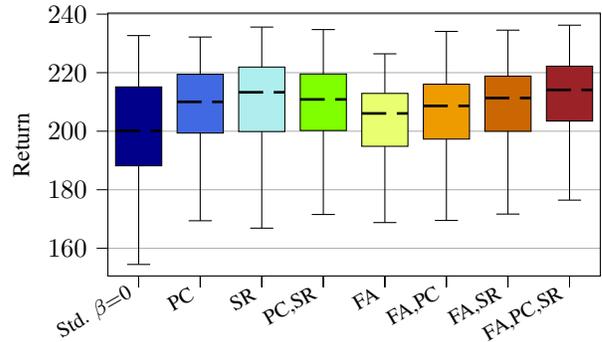
\begin{figure}[t]
  \centering
  \vspace{2mm}
\begin{tikzpicture}

\definecolor{brown1553438}{RGB}{155,34,38}
\definecolor{chartreuse}{RGB}{127,255,0}
\definecolor{chocolate2021032}{RGB}{202,103,2}
\definecolor{darkblue}{RGB}{0,0,139}
\definecolor{darkgray176}{RGB}{176,176,176}
\definecolor{khaki233255112}{RGB}{233,255,112}
\definecolor{orange2381550}{RGB}{238,155,0}
\definecolor{paleturquoise}{RGB}{175,238,238}
\definecolor{royalblue}{RGB}{65,105,225}

\begin{axis}[
height=2.0in,
tick align=outside,
tick pos=left,
width=3.2in,
x grid style={darkgray176},
xmin=0.5, xmax=8.5,
xtick style={color=black},
xtick={1,2,3,4,5,6,7,8,1,2,3,4,5,6,7,8},
xticklabel style={rotate=30.0,anchor=east, font=\footnotesize},
xticklabels={Std. $\beta$$=$0,PC,SR,{PC,SR},FA,{FA,PC},{FA,SR},{FA,PC,SR},,,,,,,,},
y grid style={darkgray176},
ylabel={\small{Return}},
ymajorgrids,
ymin=150.528806725522, ymax=240.366417344106,
ytick style={color=black}
]
\path [draw=black, fill=darkblue]
(axis cs:0.625,188.184989742728)
--(axis cs:1.375,188.184989742728)
--(axis cs:1.375,215.132744775031)
--(axis cs:0.625,215.132744775031)
--(axis cs:0.625,188.184989742728)
--cycle;
\addplot [black]
table {%
1 188.184989742728
1 154.475970844549
};
\addplot [black]
table {%
1 215.132744775031
1 232.673823729289
};
\addplot [black]
table {%
0.8125 154.475970844549
1.1875 154.475970844549
};
\addplot [black]
table {%
0.8125 232.673823729289
1.1875 232.673823729289
};
\path [draw=black, fill=royalblue]
(axis cs:1.625,199.366094568381)
--(axis cs:2.375,199.366094568381)
--(axis cs:2.375,219.449668917711)
--(axis cs:1.625,219.449668917711)
--(axis cs:1.625,199.366094568381)
--cycle;
\addplot [black]
table {%
2 199.366094568381
2 169.407939534462
};
\addplot [black]
table {%
2 219.449668917711
2 232.171919782514
};
\addplot [black]
table {%
1.8125 169.407939534462
2.1875 169.407939534462
};
\addplot [black]
table {%
1.8125 232.171919782514
2.1875 232.171919782514
};
\path [draw=black, fill=paleturquoise]
(axis cs:2.625,199.886087891061)
--(axis cs:3.375,199.886087891061)
--(axis cs:3.375,221.918487128804)
--(axis cs:2.625,221.918487128804)
--(axis cs:2.625,199.886087891061)
--cycle;
\addplot [black]
table {%
3 199.886087891061
3 166.875657760818
};
\addplot [black]
table {%
3 221.918487128804
3 235.573060754627
};
\addplot [black]
table {%
2.8125 166.875657760818
3.1875 166.875657760818
};
\addplot [black]
table {%
2.8125 235.573060754627
3.1875 235.573060754627
};
\path [draw=black, fill=chartreuse]
(axis cs:3.625,200.233551039694)
--(axis cs:4.375,200.233551039694)
--(axis cs:4.375,219.549583044452)
--(axis cs:3.625,219.549583044452)
--(axis cs:3.625,200.233551039694)
--cycle;
\addplot [black]
table {%
4 200.233551039694
4 171.545059276105
};
\addplot [black]
table {%
4 219.549583044452
4 234.730257428214
};
\addplot [black]
table {%
3.8125 171.545059276105
4.1875 171.545059276105
};
\addplot [black]
table {%
3.8125 234.730257428214
4.1875 234.730257428214
};
\path [draw=black, fill=khaki233255112]
(axis cs:4.625,194.842501172065)
--(axis cs:5.375,194.842501172065)
--(axis cs:5.375,212.919961621093)
--(axis cs:4.625,212.919961621093)
--(axis cs:4.625,194.842501172065)
--cycle;
\addplot [black]
table {%
5 194.842501172065
5 168.780649524346
};
\addplot [black]
table {%
5 212.919961621093
5 226.438310602871
};
\addplot [black]
table {%
4.8125 168.780649524346
5.1875 168.780649524346
};
\addplot [black]
table {%
4.8125 226.438310602871
5.1875 226.438310602871
};
\path [draw=black, fill=orange2381550]
(axis cs:5.625,197.306512508281)
--(axis cs:6.375,197.306512508281)
--(axis cs:6.375,216.071225178741)
--(axis cs:5.625,216.071225178741)
--(axis cs:5.625,197.306512508281)
--cycle;
\addplot [black]
table {%
6 197.306512508281
6 169.511297555809
};
\addplot [black]
table {%
6 216.071225178741
6 234.092749874851
};
\addplot [black]
table {%
5.8125 169.511297555809
6.1875 169.511297555809
};
\addplot [black]
table {%
5.8125 234.092749874851
6.1875 234.092749874851
};
\path [draw=black, fill=chocolate2021032]
(axis cs:6.625,199.966126374552)
--(axis cs:7.375,199.966126374552)
--(axis cs:7.375,218.821920330974)
--(axis cs:6.625,218.821920330974)
--(axis cs:6.625,199.966126374552)
--cycle;
\addplot [black]
table {%
7 199.966126374552
7 171.688017879457
};
\addplot [black]
table {%
7 218.821920330974
7 234.524432567706
};
\addplot [black]
table {%
6.8125 171.688017879457
7.1875 171.688017879457
};
\addplot [black]
table {%
6.8125 234.524432567706
7.1875 234.524432567706
};
\path [draw=black, fill=brown1553438]
(axis cs:7.625,203.475136778992)
--(axis cs:8.375,203.475136778992)
--(axis cs:8.375,222.214911298897)
--(axis cs:7.625,222.214911298897)
--(axis cs:7.625,203.475136778992)
--cycle;
\addplot [black]
table {%
8 203.475136778992
8 176.45147681403
};
\addplot [black]
table {%
8 222.214911298897
8 236.21929429843
};
\addplot [black]
table {%
7.8125 176.45147681403
8.1875 176.45147681403
};
\addplot [black]
table {%
7.8125 236.21929429843
8.1875 236.21929429843
};
\addplot [line width=1pt, black, dash pattern=on 9.25pt off 4pt]
table {%
0.625 200.150191781747
1.375 200.150191781747
};
\addplot [line width=1pt, black, dash pattern=on 9.25pt off 4pt]
table {%
1.625 210.034009179065
2.375 210.034009179065
};
\addplot [line width=1pt, black, dash pattern=on 9.25pt off 4pt]
table {%
2.625 213.304545330183
3.375 213.304545330183
};
\addplot [line width=1pt, black, dash pattern=on 9.25pt off 4pt]
table {%
3.625 210.863323831062
4.375 210.863323831062
};
\addplot [line width=1pt, black, dash pattern=on 9.25pt off 4pt]
table {%
4.625 206.076086988352
5.375 206.076086988352
};
\addplot [line width=1pt, black, dash pattern=on 9.25pt off 4pt]
table {%
5.625 208.649013903382
6.375 208.649013903382
};
\addplot [line width=1pt, black, dash pattern=on 9.25pt off 4pt]
table {%
6.625 211.339105877044
7.375 211.339105877044
};
\addplot [line width=1pt, black, dash pattern=on 9.25pt off 4pt]
table {%
7.625 214.107958400418
8.375 214.107958400418
};
\end{axis}

\end{tikzpicture}
  \vspace{-3mm}
  \caption{The return for the ablation study of the training modifications.}
  \label{fig:abl_reward}
  \vspace{-1mm}
\end{figure}

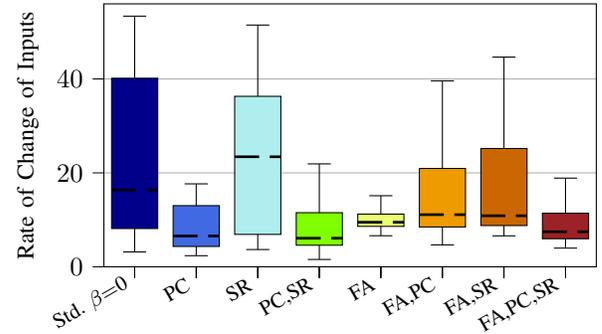
\begin{figure}[t]
  \centering
\begin{tikzpicture}

\definecolor{brown1553438}{RGB}{155,34,38}
\definecolor{chartreuse}{RGB}{127,255,0}
\definecolor{chocolate2021032}{RGB}{202,103,2}
\definecolor{darkblue}{RGB}{0,0,139}
\definecolor{darkgray176}{RGB}{176,176,176}
\definecolor{khaki233255112}{RGB}{233,255,112}
\definecolor{orange2381550}{RGB}{238,155,0}
\definecolor{paleturquoise}{RGB}{175,238,238}
\definecolor{royalblue}{RGB}{65,105,225}

\begin{axis}[
height=2.0in,
tick align=outside,
tick pos=left,
width=3.2in,
x grid style={darkgray176},
xmin=0.5, xmax=8.5,
xtick style={color=black},
xtick={1,2,3,4,5,6,7,8,1,2,3,4,5,6,7,8},
xticklabel style={rotate=30.0,anchor=east,font=\footnotesize},
xticklabels={Std. $\beta$$=$0,PC,SR,{PC,SR},FA,{FA,PC},{FA,SR},{FA,PC,SR},,,,,,,,},
y grid style={darkgray176},
ylabel={\small{Rate of Change of Inputs}},
ymajorgrids,
ymin=0, ymax=55.9819908317161,
ytick style={color=black}
]
\path [draw=black, fill=darkblue]
(axis cs:0.625,8.15070742223021)
--(axis cs:1.375,8.15070742223021)
--(axis cs:1.375,40.1454026172359)
--(axis cs:0.625,40.1454026172359)
--(axis cs:0.625,8.15070742223021)
--cycle;
\addplot [black]
table {%
1 8.15070742223021
1 3.16956852144969
};
\addplot [black]
table {%
1 40.1454026172359
1 53.3672352440883
};
\addplot [black]
table {%
0.8125 3.16956852144969
1.1875 3.16956852144969
};
\addplot [black]
table {%
0.8125 53.3672352440883
1.1875 53.3672352440883
};
\path [draw=black, fill=royalblue]
(axis cs:1.625,4.3342536947534)
--(axis cs:2.375,4.3342536947534)
--(axis cs:2.375,13.0230616513447)
--(axis cs:1.625,13.0230616513447)
--(axis cs:1.625,4.3342536947534)
--cycle;
\addplot [black]
table {%
2 4.3342536947534
2 2.3666170585322
};
\addplot [black]
table {%
2 13.0230616513447
2 17.6636393234096
};
\addplot [black]
table {%
1.8125 2.3666170585322
2.1875 2.3666170585322
};
\addplot [black]
table {%
1.8125 17.6636393234096
2.1875 17.6636393234096
};
\path [draw=black, fill=paleturquoise]
(axis cs:2.625,6.90650182149988)
--(axis cs:3.375,6.90650182149988)
--(axis cs:3.375,36.2972361521752)
--(axis cs:2.625,36.2972361521752)
--(axis cs:2.625,6.90650182149988)
--cycle;
\addplot [black]
table {%
3 6.90650182149988
3 3.6778850778078
};
\addplot [black]
table {%
3 36.2972361521752
3 51.4719303292436
};
\addplot [black]
table {%
2.8125 3.6778850778078
3.1875 3.6778850778078
};
\addplot [black]
table {%
2.8125 51.4719303292436
3.1875 51.4719303292436
};
\path [draw=black, fill=chartreuse]
(axis cs:3.625,4.59557376218258)
--(axis cs:4.375,4.59557376218258)
--(axis cs:4.375,11.5233525459857)
--(axis cs:3.625,11.5233525459857)
--(axis cs:3.625,4.59557376218258)
--cycle;
\addplot [black]
table {%
4 4.59557376218258
4 1.56499187629142
};
\addplot [black]
table {%
4 11.5233525459857
4 21.9145802125016
};
\addplot [black]
table {%
3.8125 1.56499187629142
4.1875 1.56499187629142
};
\addplot [black]
table {%
3.8125 21.9145802125016
4.1875 21.9145802125016
};
\path [draw=black, fill=khaki233255112]
(axis cs:4.625,8.61117770000352)
--(axis cs:5.375,8.61117770000352)
--(axis cs:5.375,11.2214257873872)
--(axis cs:4.625,11.2214257873872)
--(axis cs:4.625,8.61117770000352)
--cycle;
\addplot [black]
table {%
5 8.61117770000352
5 6.61459369814767
};
\addplot [black]
table {%
5 11.2214257873872
5 15.1362314872117
};
\addplot [black]
table {%
4.8125 6.61459369814767
5.1875 6.61459369814767
};
\addplot [black]
table {%
4.8125 15.1362314872117
5.1875 15.1362314872117
};
\path [draw=black, fill=orange2381550]
(axis cs:5.625,8.48073769706224)
--(axis cs:6.375,8.48073769706224)
--(axis cs:6.375,20.9427064266306)
--(axis cs:5.625,20.9427064266306)
--(axis cs:5.625,8.48073769706224)
--cycle;
\addplot [black]
table {%
6 8.48073769706224
6 4.66843106335944
};
\addplot [black]
table {%
6 20.9427064266306
6 39.5950365655648
};
\addplot [black]
table {%
5.8125 4.66843106335944
6.1875 4.66843106335944
};
\addplot [black]
table {%
5.8125 39.5950365655648
6.1875 39.5950365655648
};
\path [draw=black, fill=chocolate2021032]
(axis cs:6.625,8.78474782125691)
--(axis cs:7.375,8.78474782125691)
--(axis cs:7.375,25.1896533723708)
--(axis cs:6.625,25.1896533723708)
--(axis cs:6.625,8.78474782125691)
--cycle;
\addplot [black]
table {%
7 8.78474782125691
7 6.59613270084514
};
\addplot [black]
table {%
7 25.1896533723708
7 44.6524726650713
};
\addplot [black]
table {%
6.8125 6.59613270084514
7.1875 6.59613270084514
};
\addplot [black]
table {%
6.8125 44.6524726650713
7.1875 44.6524726650713
};
\path [draw=black, fill=brown1553438]
(axis cs:7.625,5.98252008256936)
--(axis cs:8.375,5.98252008256936)
--(axis cs:8.375,11.4077862860609)
--(axis cs:7.625,11.4077862860609)
--(axis cs:7.625,5.98252008256936)
--cycle;
\addplot [black]
table {%
8 5.98252008256936
8 3.97909157480848
};
\addplot [black]
table {%
8 11.4077862860609
8 18.8849637124403
};
\addplot [black]
table {%
7.8125 3.97909157480848
8.1875 3.97909157480848
};
\addplot [black]
table {%
7.8125 18.8849637124403
8.1875 18.8849637124403
};
\addplot [line width=1pt, black, dash pattern=on 9.25pt off 4pt]
table {%
0.625 16.3706145415124
1.375 16.3706145415124
};
\addplot [line width=1pt, black, dash pattern=on 9.25pt off 4pt]
table {%
1.625 6.56873928893962
2.375 6.56873928893962
};
\addplot [line width=1pt, black, dash pattern=on 9.25pt off 4pt]
table {%
2.625 23.4301664732217
3.375 23.4301664732217
};
\addplot [line width=1pt, black, dash pattern=on 9.25pt off 4pt]
table {%
3.625 6.08974237907204
4.375 6.08974237907204
};
\addplot [line width=1pt, black, dash pattern=on 9.25pt off 4pt]
table {%
4.625 9.49011739894848
5.375 9.49011739894848
};
\addplot [line width=1pt, black, dash pattern=on 9.25pt off 4pt]
table {%
5.625 11.1054605974388
6.375 11.1054605974388
};
\addplot [line width=1pt, black, dash pattern=on 9.25pt off 4pt]
table {%
6.625 10.8713943319679
7.375 10.8713943319679
};
\addplot [line width=1pt, black, dash pattern=on 9.25pt off 4pt]
table {%
7.625 7.4671047363877
8.375 7.4671047363877
};
\end{axis}

\end{tikzpicture}
  \vspace{-3mm}
  \caption{The rate of change of the inputs (see~\cite{multi-step}) for the ablation study of the training modifications.}
  \label{fig:abl_roc}
  \vspace{-5mm}
\end{figure}

\begin{figure}[t]
    \centering
    \vspace{3mm}
    \input{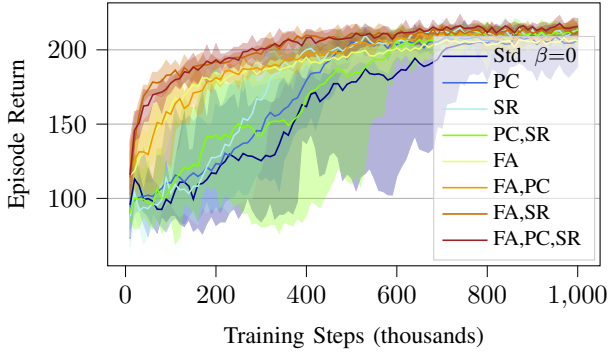}
    \vspace{-2mm}
    \caption{The return versus the number of training steps on the ablation study of the training modifications.}
    \label{fig:abl_train_return}
    \vspace{-2mm}
\end{figure}

\begin{figure}[t]
    \centering
    \input{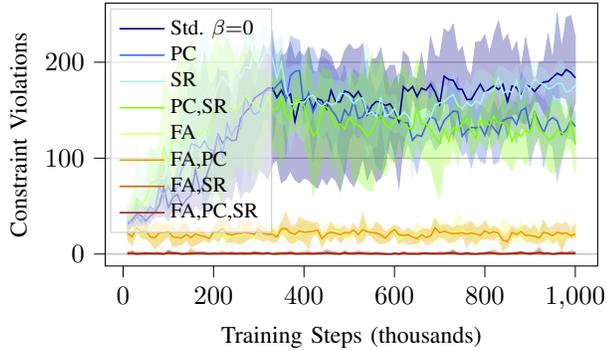}
    \vspace{-2mm}
    \caption{The number of constraint violations versus the number of training steps in the ablation study of the training modifications.}
    \label{fig:abl_viols}
    \vspace{-5mm}
\end{figure}

\begin{figure}[t]
    \centering
    \vspace{1mm}
\begin{tikzpicture}

\definecolor{brown1553438}{RGB}{155,34,38}
\definecolor{chartreuse}{RGB}{127,255,0}
\definecolor{chocolate2021032}{RGB}{202,103,2}
\definecolor{darkblue}{RGB}{0,0,139}
\definecolor{darkgray176}{RGB}{176,176,176}
\definecolor{khaki233255112}{RGB}{233,255,112}
\definecolor{orange2381550}{RGB}{238,155,0}
\definecolor{paleturquoise}{RGB}{175,238,238}
\definecolor{royalblue}{RGB}{65,105,225}

\begin{axis}[
height=2.0in,
tick align=outside,
tick pos=left,
width=3.5in,
x grid style={darkgray176},
xmin=0.5, xmax=8.5,
xtick style={color=black},
xtick={1,2,3,4,5,6,7,8,1,2,3,4,5,6,7,8},
xticklabel style={rotate=30.0,anchor=east,font=\footnotesize},
xticklabels={Std. $\beta$$=$0,PC,SR,{PC,SR},FA,{FA,PC},{FA,SR},{FA,PC,SR},,,,,,,,},
y grid style={darkgray176},
ylabel={\small{Training Time per Step [ms]}},
ymajorgrids,
ymin=0, ymax=18.8456688523293,
ytick style={color=black}
]
\path [draw=black, fill=darkblue]
(axis cs:0.625,1.88243979215622)
--(axis cs:1.375,1.88243979215622)
--(axis cs:1.375,2.43737840652466)
--(axis cs:0.625,2.43737840652466)
--(axis cs:0.625,1.88243979215622)
--cycle;
\addplot [black]
table {%
1 1.88243979215622
1 1.80297946929932
};
\addplot [black]
table {%
1 2.43737840652466
1 3.19074749946594
};
\addplot [black]
table {%
0.8125 1.80297946929932
1.1875 1.80297946929932
};
\addplot [black]
table {%
0.8125 3.19074749946594
1.1875 3.19074749946594
};
\path [draw=black, fill=royalblue]
(axis cs:1.625,9.05778169631958)
--(axis cs:2.375,9.05778169631958)
--(axis cs:2.375,10.5913544297218)
--(axis cs:1.625,10.5913544297218)
--(axis cs:1.625,9.05778169631958)
--cycle;
\addplot [black]
table {%
2 9.05778169631958
2 7.30269432067871
};
\addplot [black]
table {%
2 10.5913544297218
2 12.05672955513
};
\addplot [black]
table {%
1.8125 7.30269432067871
2.1875 7.30269432067871
};
\addplot [black]
table {%
1.8125 12.05672955513
2.1875 12.05672955513
};
\path [draw=black, fill=paleturquoise]
(axis cs:2.625,3.55293846130371)
--(axis cs:3.375,3.55293846130371)
--(axis cs:3.375,5.01835417747498)
--(axis cs:2.625,5.01835417747498)
--(axis cs:2.625,3.55293846130371)
--cycle;
\addplot [black]
table {%
3 3.55293846130371
3 2.50043964385986
};
\addplot [black]
table {%
3 5.01835417747498
3 7.1875913143158
};
\addplot [black]
table {%
2.8125 2.50043964385986
3.1875 2.50043964385986
};
\addplot [black]
table {%
2.8125 7.1875913143158
3.1875 7.1875913143158
};
\path [draw=black, fill=chartreuse]
(axis cs:3.625,10.4255487322807)
--(axis cs:4.375,10.4255487322807)
--(axis cs:4.375,13.4787120223045)
--(axis cs:3.625,13.4787120223045)
--(axis cs:3.625,10.4255487322807)
--cycle;
\addplot [black]
table {%
4 10.4255487322807
4 7.62699699401855
};
\addplot [black]
table {%
4 13.4787120223045
4 18.0341122150421
};
\addplot [black]
table {%
3.8125 7.62699699401855
4.1875 7.62699699401855
};
\addplot [black]
table {%
3.8125 18.0341122150421
4.1875 18.0341122150421
};
\path [draw=black, fill=khaki233255112]
(axis cs:4.625,10.2196543216705)
--(axis cs:5.375,10.2196543216705)
--(axis cs:5.375,10.9007322788239)
--(axis cs:4.625,10.9007322788239)
--(axis cs:4.625,10.2196543216705)
--cycle;
\addplot [black]
table {%
5 10.2196543216705
5 9.25395584106445
};
\addplot [black]
table {%
5 10.9007322788239
5 11.3910112380981
};
\addplot [black]
table {%
4.8125 9.25395584106445
5.1875 9.25395584106445
};
\addplot [black]
table {%
4.8125 11.3910112380981
5.1875 11.3910112380981
};
\path [draw=black, fill=orange2381550]
(axis cs:5.625,8.69815856218338)
--(axis cs:6.375,8.69815856218338)
--(axis cs:6.375,10.9813488125801)
--(axis cs:5.625,10.9813488125801)
--(axis cs:5.625,8.69815856218338)
--cycle;
\addplot [black]
table {%
6 8.69815856218338
6 8.01304006576538
};
\addplot [black]
table {%
6 10.9813488125801
6 11.3187339305878
};
\addplot [black]
table {%
5.8125 8.01304006576538
6.1875 8.01304006576538
};
\addplot [black]
table {%
5.8125 11.3187339305878
6.1875 11.3187339305878
};
\path [draw=black, fill=chocolate2021032]
(axis cs:6.625,10.5607879757881)
--(axis cs:7.375,10.5607879757881)
--(axis cs:7.375,12.6881381869316)
--(axis cs:6.625,12.6881381869316)
--(axis cs:6.625,10.5607879757881)
--cycle;
\addplot [black]
table {%
7 10.5607879757881
7 9.00293731689453
};
\addplot [black]
table {%
7 12.6881381869316
7 15.7656853199005
};
\addplot [black]
table {%
6.8125 9.00293731689453
7.1875 9.00293731689453
};
\addplot [black]
table {%
6.8125 15.7656853199005
7.1875 15.7656853199005
};
\path [draw=black, fill=brown1553438]
(axis cs:7.625,10.3849322795868)
--(axis cs:8.375,10.3849322795868)
--(axis cs:8.375,12.7681054472923)
--(axis cs:7.625,12.7681054472923)
--(axis cs:7.625,10.3849322795868)
--cycle;
\addplot [black]
table {%
8 10.3849322795868
8 8.85418319702148
};
\addplot [black]
table {%
8 12.7681054472923
8 16.0335237979889
};
\addplot [black]
table {%
7.8125 8.85418319702148
8.1875 8.85418319702148
};
\addplot [black]
table {%
7.8125 16.0335237979889
8.1875 16.0335237979889
};
\addplot [line width=1pt, black, dash pattern=on 9.25pt off 4pt]
table {%
0.625 2.27605378627777
1.375 2.27605378627777
};
\addplot [line width=1pt, black, dash pattern=on 9.25pt off 4pt]
table {%
1.625 9.64428508281708
2.375 9.64428508281708
};
\addplot [line width=1pt, black, dash pattern=on 9.25pt off 4pt]
table {%
2.625 4.06151056289673
3.375 4.06151056289673
};
\addplot [line width=1pt, black, dash pattern=on 9.25pt off 4pt]
table {%
3.625 11.4672068357468
4.375 11.4672068357468
};
\addplot [line width=1pt, black, dash pattern=on 9.25pt off 4pt]
table {%
4.625 10.6587266921997
5.375 10.6587266921997
};
\addplot [line width=1pt, black, dash pattern=on 9.25pt off 4pt]
table {%
5.625 10.2473353147507
6.375 10.2473353147507
};
\addplot [line width=1pt, black, dash pattern=on 9.25pt off 4pt]
table {%
6.625 11.8724406957626
7.375 11.8724406957626
};
\addplot [line width=1pt, black, dash pattern=on 9.25pt off 4pt]
table {%
7.625 11.7592972517014
8.375 11.7592972517014
};
\end{axis}

\end{tikzpicture}
    \vspace{-6mm}
    \caption{The time required to execute one training step in the ablation study of the training modifications.}
    \label{fig:abl_step_time}
    \vspace{-4mm}
\end{figure}
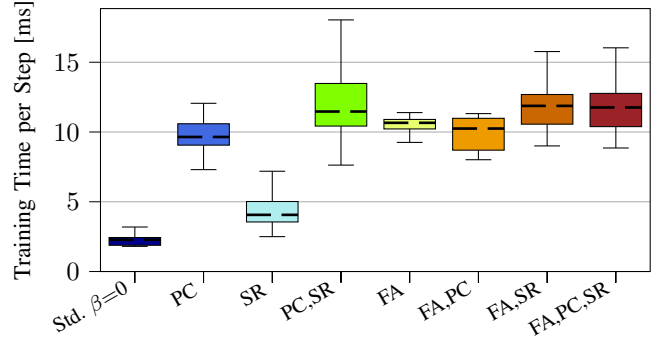

From \textref{fig:abl_viols}{Fig.}, we note that penalizing the corrections reduces the number of constraint violations during training, lowers the rate of change of the inputs (see \textref{fig:abl_roc}{Fig.}), and increases the return (see \textref{fig:abl_reward}{Fig.}), although this depends on the correction penalization weight $\alpha$ (see next section). Secondly, we find that the safe reset modification significantly improves convergence (see \textref{fig:abl_train_return}{Fig.}) and evaluation return. Finally, filtering the actions reduces the constraint violations to nearly zero, especially when partnered with the safe reset approach. Most importantly, combining all the modifications leads to the best return and convergence and the least constraint violations during training. Additionally, it results in the third lowest rate of change of the inputs.

The standard approach ``Std. $\beta$$=$0'' is significantly faster than the approaches using \diffhl{the} modifications (see \textref{fig:abl_step_time}{Fig.}), followed by the approach that only uses the safe reset modification, ``SR''. \diffhl{The} approach with all modifications ``FA,PC,SR'' is five times slower than training without any modifications. However, \diffhl{the} approaches trained with the action filtering and safe reset modifications reached 80\% of their final return ten times faster than the standard training and surpassed a return of 200 in 2.5 times fewer environment interactions (see \textref{fig:abl_train_return}{Fig.}). This sample efficiency is important when environment interactions are slow or expensive, such as training on a real system. It should also be noted that the training time depends on the type of safety filter and its implementation. A control barrier function~(CBF) based safety filter can be significantly faster than an MPSF~\cite{zeilinger_survey, fisac_survey} and could equally be used with \diffhl{the} training modifications. However, designing a CBF safety filter has other challenges, like determining the control barrier function, which remains an active area of study~\cite{brunke_safe_2021}.

\subsection{Ablation Study of Reward Penalties} \label{sec:sim_results}
To study the effects of reward penalties, we conducted experiments with various values of the correction penalization weight $\alpha$ and the constraint violation penalty $\beta$. We compare \diffhl{the safe} approach (all of \diffhl{the} training modifications together, denoted ``\diffhl{Safe}'') with $\alpha \in \{0.1, 1, 10, 100\}$ to the standard training (none of \diffhl{the} training modifications, denoted ``Std.'') with $\beta \in \{0, 0.01, 0.1, 1\}$. These experiments were performed with the same setup as the ablation study of the training modifications. The results are summarized in \textref{tbl:ablation_2}{Table}.

\begin{table*}[t]
    \addtolength{\tabcolsep}{-0.1em}
	\centering
    \vspace{2mm}
	\captionof{table}{Results for the ablation on the correction penalization weight $\alpha$ and the constraint violation penalty $\beta$.}
    \scriptsize
	\begin{tabular}{lrrrrrrrr}
		\toprule
		Metric & Std. ($\beta$$=$0) & Std. ($\beta$$=$0.01) & Std. ($\beta$$=$0.1) & Std. ($\beta$$=$1) & \diffhl{Safe} ($\alpha$$=$0.1) & \diffhl{Safe} ($\alpha$$=$1) & \diffhl{Safe} ($\alpha$$=$10) & \diffhl{Safe} ($\alpha$$=$100) \\
		\midrule
		Return                                        & 200.2 $\pm$ 17.1 & 208.3 $\pm$ 15.5 & 210.3 $\pm$ 14.4 & 199.1 $\pm$ 15.6 & 212.6 $\pm$ 13.8 & \textbf{214.1 $\pm$ 14.0} & 210.7 $\pm$ 14.0 & 202.4 $\pm$ 13.4 \\
		Return when uncertified                      & 222.1 $\pm$ 13.7 & \textbf{229.0 $\pm$ 10.1} & 222.2 $\pm$ 10.9 & 206.8 $\pm$ 14.6 & 11.2 $\pm$ 15.7 & 31.2 $\pm$ 44.8 & 210.7 $\pm$ 50.0 & 202.3 $\pm$ 19.1 \\
        Input rate of change [\si{\meter\per\second}] & 16.4 $\pm$ 17.0  & 10.0 $\pm$ 13.8 & 4.7 $\pm$ 2.9\hpz & 3.2 $\pm$ 3.0\hpz & 9.4 $\pm$ 2.9\hpz & 7.5 $\pm$ 3.3\hpz & 3.8 $\pm$ 3.0\hpz & \textbf{3.0 $\pm$ 3.1}\hpz \\
		Training constraint violations   [\%]         & 82.8 $\pm$ 6.6\hpz & 78.0 $\pm$ 5.4\hpz & 67.0 $\pm$ 4.2\hpz & 74.5 $\pm$ 9.9\hpz & 0.23 $\pm$ 0.02 & 0.22 $\pm$ 0.03 & 0.22 $\pm$ 0.02 & \textbf{0.22 $\pm$ 0.02} \\
        Training time per step [\si{\milli\second}]               & \textbf{2.3 $\pm$ 0.3}\hpz & 2.7 $\pm$ 0.2\hpz & 2.8 $\pm$ 0.4\hpz & 2.8 $\pm$ 0.2\hpz & 14.6 $\pm$ 1.2\hpz & 11.6 $\pm$ 1.5\hpz & 13.5 $\pm$ 1.0\hpz & 12.6 $\pm$ 1.1\hpz \\
		\bottomrule
	\end{tabular}
	\label{tbl:ablation_2}
	\vspace*{-5mm}
\end{table*}

In the experiments (evaluated with the safety filter), we note that as $\alpha$ and $\beta$ increase, the rate of change of the inputs decreases. The return for \diffhl{the safe} approach is maximized by $\alpha=1$, while the return is maximized for the baseline with $\beta=0.1$. The baseline did not converge with $\beta > 1$.

\diffhl{The safe} approach increases the return by up to 7.3\% compared to the standard approaches with $\beta=0$, and up to 2.0\% compared to the best performing standard approach, $\beta=0.1$. The rate of change of the inputs is reduced by up to 75\% compared to the baseline with $\beta=0$, and up to 7.8\% compared to the smoothest baseline, $\beta=1$. This demonstrates that \diffhl{the safe} training approach effectively reduces chattering and jerky motions due to undesirable interplay between the safety filter and the controller and increases the controller's performance when certified.

\begin{figure}[t]
    \centering
    \input{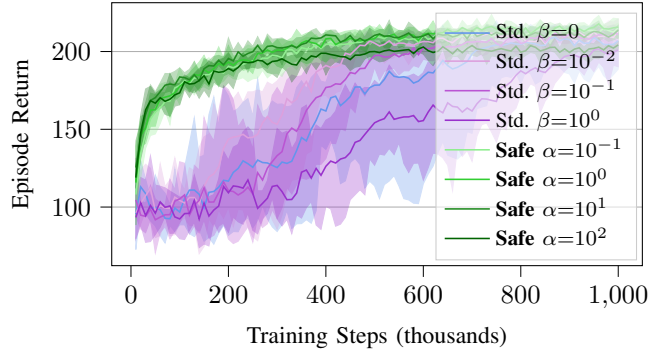}
    \vspace{-6mm}
    \caption{The return versus the number of training steps on the ablation study of reward penalties.}
    \label{fig:return}
    \vspace{-6mm}
\end{figure}

Further, let us consider the convergence of \diffhl{the safe} approach. From \textref{fig:return}{Fig.}, \diffhl{the} fastest converging \diffhl{safe} approach (``\diffhl{Safe} $\alpha=10$'') reached 80\% of its final performance seven times faster than the fastest converging standard training (``Std. $\beta=0.01$'') and surpassed a return of 200 with 1.8 times fewer environment interactions. We found that as the complexity and difficulty in ensuring the safety of the experiment increased, the benefit of \diffhl{the safe} approach in terms of convergence increased, and the convergence of the baseline approaches slowed significantly. Additionally, from \textref{tbl:ablation_2}{Table}, when using \diffhl{the safe} approach, only 0.22\% of the training steps violate constraints. In contrast, the baseline with $\beta=0$ and $\beta=0.1$ (the $\beta$ value resulting in the least constraint violations) violate the constraints 82.8\% and 67.0\% of the time, respectively. \diffhl{The safe} approaches' small number of constraint violations is due to softening the constraints in the MPSF formulation, which improved training efficiency. There were no constraint violations during the evaluation. 

\begin{figure}[t]
  \centering
  \vspace{2mm}
\begin{tikzpicture}

\definecolor{lightgray204}{RGB}{204,204,204}
\definecolor{darkgray176}{RGB}{176,176,176}
\definecolor{limegreen}{RGB}{50,205,50}
\definecolor{lightgreen}{RGB}{144,238,144}
\definecolor{forestgreen}{RGB}{34,139,34}
\definecolor{darkgreen}{RGB}{0,100,0}
\definecolor{cornflowerblue}{RGB}{100,149,237}
\definecolor{plum}{RGB}{221,160,221}
\definecolor{mediumorchid}{RGB}{186,85,211}
\definecolor{darkorchid}{RGB}{153,50,204}

\begin{axis}[
height=2.0in,
legend cell align={left},
legend style={
  fill opacity=0.6,
  draw opacity=1,
  text opacity=1,
  at={(0.01,0.98)},
  anchor=north west,
  draw=lightgray204,
  font=\footnotesize,
  row sep=-1,
},
tick align=outside,
tick pos=left,
width=3.3in,
x grid style={darkgray176},
xmin=0.3, xmax=8.7,
xtick style={color=black},
xtick={1,2,3,4,5,6,7,8},
xticklabel style={rotate=30.0,anchor=east, font=\footnotesize},
xticklabels={
  Std. $\beta$$=$$0$,
  Std. $\beta$$=$$10^{-2}$,
  Std. $\beta$$=$$10^{-1}$,
  Std. $\beta$$=$$10^{0}$,
  \textbf{\diffhl{Safe} $\boldsymbol{\alpha$$=$$10^{-1}}$},
  \textbf{\diffhl{Safe} $\boldsymbol{\alpha$$=$$10^0}$},
  \textbf{\diffhl{Safe} $\boldsymbol{\alpha$$=$$10^1}$},
  \textbf{\diffhl{Safe} $\boldsymbol{\alpha$$=$$10^2}$},
},
y grid style={darkgray176},
ylabel={\small{Constraint Violations}},
ymajorgrids,
ymin=-1, ymax=262.5,
ytick style={color=black}
]
\path [draw=black, fill=cornflowerblue, postaction={pattern=north west lines}]
(axis cs:0.625,145)
--(axis cs:1.375,145)
--(axis cs:1.375,201)
--(axis cs:0.625,201)
--(axis cs:0.625,145)
--cycle;
\addplot [black]
table {%
1 145
1 85
};
\addplot [black]
table {%
1 201
1 250
};
\addplot [black]
table {%
0.8125 85
1.1875 85
};
\addplot [black]
table {%
0.8125 250
1.1875 250
};
\path [draw=black, fill=plum, postaction={pattern=north west lines}]
(axis cs:1.625,108)
--(axis cs:2.375,108)
--(axis cs:2.375,185)
--(axis cs:1.625,185)
--(axis cs:1.625,108)
--cycle;
\addplot [black]
table {%
2 108
2 51
};
\addplot [black]
table {%
2 185
2 208
};
\addplot [black]
table {%
1.8125 51
2.1875 51
};
\addplot [black]
table {%
1.8125 208
2.1875 208
};
\path [draw=black, fill=mediumorchid, postaction={pattern=north west lines}]
(axis cs:2.625,48)
--(axis cs:3.375,48)
--(axis cs:3.375,96)
--(axis cs:2.625,96)
--(axis cs:2.625,48)
--cycle;
\addplot [black]
table {%
3 48
3 0
};
\addplot [black]
table {%
3 96
3 157
};
\addplot [black]
table {%
2.8125 0
3.1875 0
};
\addplot [black]
table {%
2.8125 157
3.1875 157
};
\path [draw=black, fill=darkorchid, postaction={pattern=north west lines}]
(axis cs:3.625,8)
--(axis cs:4.375,8)
--(axis cs:4.375,48)
--(axis cs:3.625,48)
--(axis cs:3.625,8)
--cycle;
\addplot [black]
table {%
4 8
4 0
};
\addplot [black]
table {%
4 48
4 106
};
\addplot [black]
table {%
3.8125 0
4.1875 0
};
\addplot [black]
table {%
3.8125 106
4.1875 106
};
\path [draw=black, fill=lightgreen, postaction={pattern=north west lines}]
(axis cs:4.625,250)
--(axis cs:5.375,250)
--(axis cs:5.375,250)
--(axis cs:4.625,250)
--(axis cs:4.625,250)
--cycle;
\addplot [black]
table {%
5 250
5 250
};
\addplot [black]
table {%
5 250
5 250
};
\addplot [black]
table {%
4.8125 250
5.1875 250
};
\addplot [black]
table {%
4.8125 250
5.1875 250
};
\path [draw=black, fill=limegreen, postaction={pattern=north west lines}]
(axis cs:5.625,248)
--(axis cs:6.375,248)
--(axis cs:6.375,250)
--(axis cs:5.625,250)
--(axis cs:5.625,248)
--cycle;
\addplot [black]
table {%
6 248
6 245
};
\addplot [black]
table {%
6 250
6 250
};
\addplot [black]
table {%
5.8125 245
6.1875 245
};
\addplot [black]
table {%
5.8125 250
6.1875 250
};
\path [draw=black, fill=forestgreen, postaction={pattern=north west lines}]
(axis cs:6.625,41.75)
--(axis cs:7.375,41.75)
--(axis cs:7.375,129.5)
--(axis cs:6.625,129.5)
--(axis cs:6.625,41.75)
--cycle;
\addplot [black]
table {%
7 41.75
7 0
};
\addplot [black]
table {%
7 129.5
7 250
};
\addplot [black]
table {%
6.8125 0
7.1875 0
};
\addplot [black]
table {%
6.8125 250
7.1875 250
};
\path [draw=black, fill=darkgreen, postaction={pattern=north west lines}]
(axis cs:7.625,9)
--(axis cs:8.375,9)
--(axis cs:8.375,93.25)
--(axis cs:7.625,93.25)
--(axis cs:7.625,9)
--cycle;
\addplot [black]
table {%
8 9
8 0
};
\addplot [black]
table {%
8 93.25
8 161
};
\addplot [black]
table {%
7.8125 0
8.1875 0
};
\addplot [black]
table {%
7.8125 161
8.1875 161
};
\addplot [line width=1pt, black, dash pattern=on 9.25pt off 4pt]
table {%
0.625 176
1.375 176
};
\addplot [line width=1pt, black, dash pattern=on 9.25pt off 4pt]
table {%
1.625 134
2.375 134
};
\addplot [line width=1pt, black, dash pattern=on 9.25pt off 4pt]
table {%
2.625 75.5
3.375 75.5
};
\addplot [line width=1pt, black, dash pattern=on 9.25pt off 4pt]
table {%
3.625 23
4.375 23
};
\addplot [line width=1pt, black, dash pattern=on 9.25pt off 4pt]
table {%
4.625 250
5.375 250
};
\addplot [line width=1pt, black, dash pattern=on 9.25pt off 4pt]
table {%
5.625 250
6.375 250
};
\addplot [line width=1pt, black, dash pattern=on 9.25pt off 4pt]
table {%
6.625 96
7.375 96
};
\addplot [line width=1pt, black, dash pattern=on 9.25pt off 4pt]
table {%
7.625 38.5
8.375 38.5
};
\end{axis}

\end{tikzpicture}
  \vspace{-2mm}
  \caption{The number of constraint violations during evaluation when not using a safety filter for the ablation study of reward penalties. With the safety filter, there are no constraint violations.}
  \label{fig:eval_viols}
  \vspace{-6mm}
\end{figure}
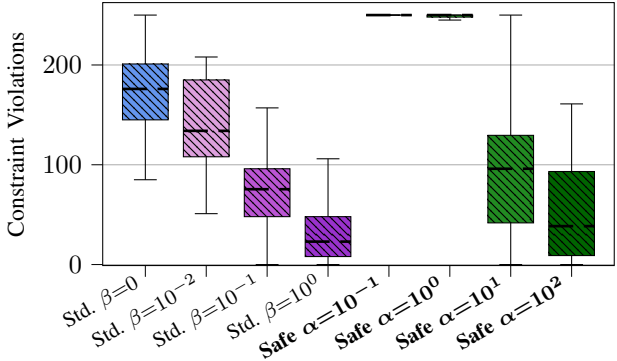

When the safety filter is removed during evaluation, all the approaches violate the constraints (see \textref{fig:eval_viols}{Fig.}), although increasing $\alpha$ or $\beta$ leads to fewer constraint violations. The return of approaches trained without the safety filter is reduced with the addition of the safety filter, demonstrating that they do not operate optimally when the safety filter is applied in evaluation (see ``Return when uncertified'' in \textref{tbl:ablation_2}{Table}). However, the approaches trained with the safety filter have the same or better return with the safety filter. This highlights the importance of not removing the safety filter, even when penalizing corrections, as safety guarantees are lost and performance may decrease.

\subsection{Real Experiments} \label{sec:real_results}
\diffhl{The safe} approach was tested on a real Crazyflie 2.0 quadrotor with a desired trajectory of a figure-eight in the $x-y$ plane with an amplitude of \SI{2}{\meter}, which must be completed twice in \SI{15}{\second}. The controllers were trained in a simulation of the real Crazyflie based on the sim-to-real stack from~\cite{sim2real}, simulating both its dynamics and internal software. Additionally, the controllers were trained with domain randomization and added Gaussian noise to the dynamics and observations to improve their ability to generalize from the simulated environment to the real one. The trained controllers were then evaluated on the real Crazyflies.

We consider the state $\textbf{x} = [x, \dot{x}, y, \dot{y}, \phi, \theta]\T \in \RR^6$. The $x$ and $y$ positions of the drone are constrained to be within [\SI{-0.95}{\meter}, \SI{0.95}{\meter}], the velocities of the drone in $x$ and $y$ are constrained to be within [\SI{-2}{\meter\per\second}, \SI{2}{\meter\per\second}], and the roll $\phi$ and pitch $\theta$ angles are constrained to be within [\SI{-0.25}{\radian}, \SI{0.25}{\radian}]. The input $\textbf{u}  = [\phi, \theta]\T \in \RR^2$ are the desired roll and pitch angles which the quadrotor tracks using an internal PID controller. The commanded angles are constrained to be within [\SI{-0.25}{\radian}, \SI{0.25}{\radian}]. The closed-loop dynamics in the $x-y$ plane can be approximated as a linear system. The linear system was identified from the uncertified flights of the baseline approaches, using simulation data during training and real data during real evaluation.

The maximum model mismatch was significantly higher than for the simulation experiments, with $w_{\text{max}} = 0.1267$. However, there were a significant number of outliers, so rather than using the maximum error, we used the mean error plus three standard deviations, $w_{\sigma=3} = 0.0515$. A motion capture system measures the quadrotor pose during the real experiment. 

\diffhl{The safe} approaches (all of \diffhl{the} training modifications together, denoted ``\diffhl{Safe}'') with $\alpha \in \{0.1, 1, 10\}$ and the baselines (none of \diffhl{the} training modifications, denoted ``Std.'')  with $\beta \in \{0, 1\}$ are tested five times each with an MPC horizon of $H=10$ and a control frequency of \SI{25}{\hertz}. The total number of iterations, and thus the maximum return, is \SI{25}{\hertz} $\cdot$ \SI{15}{\second} $= 375$. Every test starts at the same initial position, the origin of the $x-y$ plane, and a height of \SI{1}{\meter}. 

As seen in \textref{tbl:real_experiments}{Table}, \diffhl{the safe} approach reduces the rate of change of the inputs by up to 35\% compared to the $\beta=0$ baseline and increases the total return by up to 20\%. Compared to the $\beta=1$ baseline, the return is increased by up to 1.5\%, and the norm of the rate of change of the inputs is increased by 7\%. This is likely due to the simplified nature of the real experiments compared to the simulation experiments and the conservativeness of the $\beta=1$ baseline. Interestingly, \diffhl{the safe} approach provides a greater proportional increase in the return of real experiments compared to simulation. This may be due to the increased conservativeness of the safety filter, which causes a larger distribution shift between the certified and uncertified environments.

Once again, \diffhl{the safe} approach significantly decreases constraint violations in training despite softened constraints, domain randomization, and added Gaussian noise. We reduce constraint violations by over 17 times compared to the $\beta=0$ baseline and five times compared to the $\beta=1$ baseline. \diffhl{The} baseline approaches \diffhl{execute training steps} 13\% faster than \diffhl{the safe} approaches, compared to five times faster in the simulation sections (see \textref{fig:abl_step_time}{Fig.} and \textref{tbl:ablation_2}{Table}). This is because the simulation used in this section includes the internal software of the Crazyflies~\cite{sim2real}, significantly increasing the time per training step. This highlights the importance of improving sample efficiency. 

\begin{table*}[t]
	\centering
    \vspace{2mm}
	\captionof{table}{Results for the real experiments of five trials flying a Crazyflie 2.0 on a figure-eight path.}
    \arrayrulecolor{black}
	\begin{tabular}{lrrrrrr}
		\toprule
		Metric & Std. ($\beta$$=$$0$) & Std. ($\beta$$=$1) & \diffhl{Safe} ($\alpha$$=$$0.1$) & \diffhl{Safe} ($\alpha$$=$$1$) & \diffhl{Safe} ($\alpha$$=$$10$) \\
		\midrule
		Return                                                & 263.5 $\pm$ 7.7\hpz & 311.7 $\pm$ 4.4\hpz & 314.2 $\pm$ 9.7\hpz & \textbf{316.1 $\pm$ 1.4}\hpz & 305.5 $\pm$ 0.8\hpz \\
        Input rate of change [\si{\meter\per\second}]         & 237.3 $\pm$ 15.9 & \textbf{143.8 $\pm$ 5.0}\hpz& 234.8 $\pm$ 13.0 & 199.9 $\pm$ 6.0\hpz& 154.0 $\pm$ 8.7\hpz \\
		Training constraint violations [\%]                   & 47.7 $\pm$ 14.6 & 14.3 $\pm$ 12.8 & 11.9 $\pm$ 10.9 & 6.3 $\pm$ 7.5\hpz & \textbf{2.7 $\pm$ 3.0}\hpz  \\
        Training time per step [\si{\milli\second}]  & 14.4 $\pm$ 0.1\hpz & \textbf{14.2 $\pm$ 0.1}\hpz & 16.4 $\pm$ 0.1\hpz & 16.4 $\pm$ 0.1\hpz & 16.1 $\pm$ 0.1\hpz \\
		\bottomrule
	\end{tabular}
	\label{tbl:real_experiments}
	\vspace*{-4mm}
\end{table*}

\section{Conclusion}
This paper \diffhl{analyzes} three methods of incorporating safety filters in the training of model-free RL algorithms: filtering the training actions, penalizing the reward by the magnitude of the corrections, and resetting episodes to safe starting states. The presented modifications require little tuning and have been designed to be easily incorporated into the training of any RL algorithm. In simulation and real experiments done using model predictive safety filters, we found that the training modifications improve performance and sample efficiency, reduce chattering, and nearly eliminate training-time constraint violations. These modifications further leverage safety filters, allowing RL algorithms to be effectively applied to safety-critical systems.

\balance
\bibliographystyle{IEEEtran} 
\bibliography{bibliography}
\balance
\end{document}